# Unpredictability of AI

**Roman V. Yampolskiy**
Computer Engineering and Computer Science
University of Louisville
roman.yampolskiy@louisville.edu, @romanyam
May 29, 2019

> *"As machines learn they may develop unforeseen strategies at rates that baffle their programmers"*
> Norbert Wiener, 1960

> *"It's a problem writers face every time we consider the creation of intelligences greater than our own."*
> Vernor Vinge, 2001

> *"The creative unpredictability of intelligence is not like the noisy unpredictability of a random number generator."*
> Eliezer Yudkowsky, 2008

**Abstract**
The young field of AI Safety is still in the process of identifying its challenges and limitations. In this paper, we formally describe one such impossibility result, namely Unpredictability of AI. We prove that it is impossible to precisely and consistently predict what specific actions a smarter-than-human intelligent system will take to achieve its objectives, even if we know terminal goals of the system. In conclusion, impact of Unpredictability on AI Safety is discussed.

## 1. Introduction to Unpredictability

With increase in capabilities of artificial intelligence, over the last decade, a significant number of researchers have realized importance in creating not only capable intelligent systems, but also making them safe and secure [1-6]. Unfortunately, the field of AI Safety is very young, and researchers are still working to identify its main challenges and limitations. Impossibility results are well known in many fields of inquiry [7-13], and some have now been identified in AI Safety [14-16]. In this paper, we concentrate on a poorly understood concept of unpredictability of intelligent systems [17], which limits our ability to understand impact of intelligent systems we are developing and is a challenge for software verification and intelligent system control, as well as AI Safety in general.

In theoretical computer science and in software development in general, many well-known impossibility results are well established, some of them are strongly related to the subject of this paper, for example: Rice's Theorem states that no computationally effective method can decide if a program will exhibit a particular non-trivial behavior, such as producing a specific output [18]. Similarly, Wolfram's Computational Irreducibility states that complex behaviors of programs can't be predicted without actually running those programs [19]. Any physical system which could be mapped onto a Turing Machine will similarly, exhibit Unpredictability [20, 21].

*Unpredictability* of AI, one of many impossibility results in AI Safety also known as Unknowability [22] or Cognitive Uncontainability [23], is defined as our inability to precisely and consistently predict what specific actions an intelligent system will take to achieve its objectives, even if we know terminal goals of the system. It is related but is not the same as unexplainability and incomprehensibility of AI. Unpredictability does not imply that better-than-random statistical analysis is impossible; it simply points out a general limitation on how well such efforts can perform, and is particularly pronounced with advanced generally intelligent systems (superintelligence) in novel domains. In fact we can present a proof of unpredictability for such, superintelligent, systems.

*Proof*. This is a proof by contradiction. Suppose not, suppose that unpredictability is wrong and it is possible for a person to accurately predict decisions of superintelligence. That means they can make the same decisions as the superintelligence, which makes them as smart as superintelligence but that is a contradiction as superintelligence is defined as a system smarter than any person is. That means that our initial assumption was false and unpredictability is not wrong. ∎

The amount of unpredictability can be formally measured via the theory of Bayesian surprise, which measures the difference between posterior and prior beliefs of the predicting agent [24-27]. "The unpredictability of intelligence is a very special and unusual kind of surprise, which is not at all like noise or randomness. There is a weird balance between the unpredictability of actions and the predictability of outcomes." [28]. A simple heuristic is to estimate the amount of surprise as proportionate to the difference in intelligence between the predictor and the predicted agent. See Yudkowsky [29, 30] for an easy to follow discussion on this topic.

Unpredictability is practically observable in current narrow domain systems with superhuman performance. Developers of famous intelligent systems such as Deep Blue (Chess) [31, 32], IBM Watson (Jeopardy) [33], and AlphaZero (Go) [34, 35] did not know what specific decisions their AI is going to make for every turn. All they could predict was that it would try to win using any actions available to it, and win it did. AGI developers are in exactly the same situation, they may know the ultimate goals of their system but do not know the actual step-by-step plan it will execute, which of course has serious consequences for AI Safety [36-39]. A reader interested in concrete examples of unanticipated actions of intelligent agents is advised to read two surveys on the subject, one in the domain of evolutionary algorithms [40] and another on narrow AI agents [41].

There are infinitely many paths to every desirable state of the world. Great majority of them are completely undesirable and unsafe, most with negative side effects. In harder and most real-world cases, even the overall goal of the system may not be precisely know or may be known only in abstract terms, aka "make the world better". While in some cases the terminal goal(s) could be learned, even if you can learn to predict an overall outcome with some statistical certainty, you cannot learn to predict all the steps to the goal a system of superior intelligence would take. Lower intelligence can't accurately predict all decisions of higher intelligence, a concept known as *Vinge's Principle* [42]. "Vinge's Principle implies that when an agent is designing another agent (or modifying its own code), it needs to approve the other agent's design without knowing the other agent's exact future actions." [43].

## 2. Predictability: What We Can Predict – A Literature Review

Early realization that superintelligent machines will produce unpredictable futures can be found in the seminal paper by Vernor Vinge [22] on Technological Singularity, in which he talks about unknowable prediction horizon (see also Event Horizon Thesis [44]) beyond which we can predict nothing: "Perhaps it was the science-fiction writers who felt the first concrete impact. After all, the hard science-fiction writers are the ones who try to write specific stories about all that technology may do for us. More and more, these writers felt an opaque wall across the future. Once, they could put such fantasies millions of years in the future. Now they saw that their most diligent extrapolations resulted in the unknowable …" [22]. However, not everyone agreed [45]. In this section, we provide examples from literature discussing what properties of intelligent systems may in fact be predictable.

Nick Bostrom in his comment, *Singularity and Predictability,* on Vinge's work says [46]: "I am not at all sure that Unpredictability would hold. … I think there are some things that we can predict with a reasonable degree of confidence beyond the singularity. For example, that the superintelligent entity resulting from the singularity would start a spherical colonization wave that would propagate into space at a substantial fraction of the speed of light. … Another example is that if there are multiple independent competing agents (which I suspect there might not be) then we would expect some aspects of their behaviour to be predictable from considerations of economic rationality. … It might also be possible to predict things in much greater detail. Since the superintelligences or posthumans that will govern the post-singularity world will be created by us, or might even *be* us, it seems that we should be in a position to influence what values they will have. What their values are will then determine what the world will look like, since due to their advanced technology they will have a great ability to make the world conform to their values and desires. So one could argue that all we have to do in order to predict what will happen after the singularity is to figure out what values the people will have who will create the superintelligence. ... So maybe we can define a fairly small number of hypothesis about what the post-singularity world will be like. Each of these hypotheses would correspond to a plausible value. The plausible values are those that it seems fairly probable that many of the most influential people will have at about the time when the first superintelligence is created. Each of these values defines an attractor, i.e. a state of the world which contains the maximal amount of positive utility according to the value in question. We can then make the prediction that the world is likely to settle into one of these attractors. More specifically, we would expect that within the volume of space that has been colonized, matter would gradually (but perhaps very quickly) be arranged into value-maximal structures -- structures that contain as much of the chosen value as possible." [46].

Similarly, Michael Nielsen argues [47]: "What does "unknowable" mean? It seems to me that the sense in which Vinge uses the term unknowable is equivalent to "unpredictable", so let's ask the question "Will the future after the advent of dominant AI necessarily be unpredictable?" instead. … It seems to me to be ridiculous to claim that we can't make useful predictions about a post-Dominant AI world. Yes, things will change enormously. Our predictions may be much less reliable than before. However, I believe that we can still make some reasonable predictions about such a future. At the very least, we can work on excluding some possibilities. One avenue for doing this is to look at exclusions based upon the laws of physics. An often-made assertion related to the "unpredictability" of a post-Dominant-AI future is that anything allowed by the Laws of Physics will become possible at that point." [47].

Arbital contributors in their discussion of Vingean Uncertainty write: "Furthermore, our ability to think about agents smarter than ourselves is not limited to knowing a particular goal and predicting its achievement. If we found a giant alien machine that seemed very well-designed, we might be able to infer the aliens were superhumanly intelligent even if we didn't know the aliens' ultimate goals. If we saw metal pipes, we could guess that the pipes represented some stable, optimal mechanical solution which was made out of hard metal so as to retain its shape. If we saw superconducting cables, we could guess that this was a way of efficiently transporting electrical work from one place to another, even if we didn't know what final purpose the electricity was being used for. This is the idea behind Instrumental convergence: if we can recognize that an alien machine is efficiently harvesting and distributing energy, we might recognize it as an intelligently designed artifact in the service of *some* goal even if we don't know the goal." [31].

""Vingean uncertainty" is the peculiar epistemic state we enter when we're considering sufficiently intelligent programs; in particular, we become less confident that we can predict their exact actions, and more confident of the final outcome of those actions. (Note that this rejects the claim that we are epistemically helpless and can know nothing about beings smarter than ourselves.)" [31]. Yudkowsky and Herreshoff reiterate: "Hence although we cannot predict the exact actions of a smarter agent, we may be able to predict the consequences of running that agent by inspecting its design, or select among possible consequences by selecting among possible designs." [48].

Arguments against unpredictability are usually of two general types: "The apparent knownness of a particular domain. (E.g., since we have observed the rules of chemistry with great precision and know their origin in the underlying molecular dynamics, we can believe that even an arbitrarily smart agent should not be able to turn lead into gold using non-radioactive chemical reagents.) … Backward reasoning from the Fermi Paradox, which gives us weak evidence bounding the capabilities of the most powerful agents possible in our universe. (E.g., even though there might be surprises remaining in the question of how to standardly model physics, any surprise yielding Faster-Than-Light travel to a previously un-traveled point makes the Fermi Paradox harder to explain.)" [49].

In a more practical demonstration of predictability, Israeli and Goldenfeld "… find that computationally irreducible physical processes can be predictable and even computationally reducible at a coarse-grained level of description. The resulting coarse-grained [cellular automata] which we construct emulate the large-scale behavior of the original systems without accounting for small-scale details." [50]. Much of the future work on AI Safety will have to deal with figuring out what is in fact predictable and knowable about intelligent machines, even if most of their future states will forever be unpredictable to us. Next section looks at the early progress in that effort.

## 3. Cognitive Uncontainability

Machine Intelligence Research Institute (MIRI), a leading AI Safety research organization, in the context of their work on safe self-improvement in artificially intelligent agents has investigated unpredictability under the designation of *Cognitive Uncontainability*. The term is meant to infer that the human mind does not and could not conceive of all possible decisions/strategies such advanced intelligent systems could make. "Strong cognitive uncontainability is when the agent knows some facts we don't, that it can use to formulate strategies that we wouldn't be able to recognize in advance as successful. … When an agent can win using options that we didn't imagine, couldn't invent, and wouldn't understand even if we caught a glimpse of them in advance, it is strongly cognitively uncontainable …" [23]. "[I]f extremely high confidence must be placed on the ability of self-modifying systems to reason about agents which are smarter than the reasoner, then it seems prudent to develop a theoretical understanding of satisfactory reasoning about smarter agents." [51].

Even subhuman in general sense narrow domain AIs may be unpredictable to human researchers. "Although Vingean unpredictability is the classic way in which cognitive uncontainability can arise, other possibilities are imaginable. For instance, the AI could be operating in a rich domain and searching a different part of the search space that humans have difficulty handling, while still being dumber or less competent overall than a human. In this case the AI's strategies might still be unpredictable to us, even while it was less effective or competent overall." [23].

"Arguments in favor of strong uncontainability tend to revolve around either:

- The richness and partial unknownness of a particular domain. (E.g. human psychology seems very complicated; has a lot of unknown pathways; and previously discovered exploits often seemed very surprising; therefore we should expect strong uncontainability on the domain of human psychology.)
- Outside-view induction on previous ability advantages derived from cognitive advantages. (The 10th century couldn't contain the 20th century even though all parties involved were biological Homo sapiens; what makes us think we're the first generation to have the real true laws of the universe in our minds?)" [49].

## 4. Conclusions

Unpredictability is an intuitively familiar concept, we can usually predict outcome of common physical processes without knowing specific behavior of particular atoms, just like we can typically predict overall behavior of the intelligent system without knowing specific intermediate steps. Rahwan and Cebrian observe that "… complex AI agents often exhibit inherent unpredictability: they demonstrate emergent behaviors that are impossible to predict with precision—even by their own programmers. These behaviors manifest themselves only through interaction with the world and with other agents in the environment. … In fact, Alan Turing and Alonzo Church showed the fundamental impossibility of ensuring an algorithm fulfills certain properties without actually running said algorithm. There are fundamental theoretical limits to our ability to verify that a particular piece of code will always satisfy desirable properties, unless we execute the code, and observe its behavior." [52]. See Rahwan et al. for additional discussion on unpredictability and related issues with machine behavior [53].

Others have arrived at similar conclusions. "Given the inherent unpredictability of AI, it may not always be feasible to implement specific controls for every activity in which a bot engages." [54]. "As computer programs become more intelligent and less transparent, not only are the harmful effects less predictable, but their decision-making process may also be unpredictable." [55]. "The AI could become so complex that it results in errors and unpredictability, as the AI will be not able to predict its own behavior." [56]. "… the behavior of [artificial intellects] will be so complex as to be unpredictable, and therefore potentially threatening to human beings." [57].

In the context of AI safety [58-61] and AI governance [62] unpredictability implies that certain standard tools and safety mechanisms would not work to make advanced intelligent systems safe to use. For example, Bathaee writes about legislative control: "… unpredictability makes it very unlikely that the law can appropriately encourage or deter certain effects, and more problematically, the failure of our legal structures will allow people using the algorithms to externalize costs to others without having the ability to pay for the injuries they inflict." [55].

We can conclude that Unpredictability of AI will forever make 100% safe AI an impossibility, but we can still strive for Safer AI, because we are able to make some predictions about AIs we design. Terminal goals of agents can be completely arbitrary [63], but instrumental goals [64] are universal and all sufficiently intelligent agents will converge on them. Additional analysis indicated that all instrumental goals could be reduced to just one single drive, to become the most intelligent agent possible, a meta-goal for all agents, which as a side-effect may produce superconsciousness [65] impact of which on agent's behavior make may make them even less predictable to us.

## Acknowledgments
The author is grateful to Elon Musk and the Future of Life Institute, and to Jaan Tallinn and Effective Altruism Ventures for partially funding his work on AI Safety. Impact from this paper on behavior of future superintelligent machines can't be predicted.